\documentclass[letterpaper, 10 pt, conference]{ieeeconf}  

\IEEEoverridecommandlockouts                              

\usepackage{graphics} 
\usepackage{amsmath} 
\usepackage{multirow} 
\usepackage{float}
\usepackage{booktabs} 
\usepackage{epsfig} 
\usepackage{textgreek}
\usepackage{siunitx}

\makeatletter
\let\NAT@parse\undefined
\makeatother
\usepackage[colorlinks,
            linkcolor=black,
            anchorcolor=black, 
            citecolor=black]{hyperref} 

\title{\LARGE \bf
Portable, High-Frequency, and High-Voltage Control Circuits for Untethered Miniature Robots Driven by Dielectric Elastomer Actuators}

\author{Qi Shao, Xin-Jun Liu, and Huichan Zhao$^*$, \IEEEmembership{Member,~IEEE} 
\thanks{This work was supported by the National Natural Science Foundation of China under Grant Nos. 52222502 and 92048302, Beijing Municipal Natural Science Foundation under Grant No. E2024202287, and the National Key R\&D Program of China under Grant No. 2023YFB4704700. (*\textit{Corresponding author: Huichan Zhao}, {\tt\small zhaohuichan@mail.tsinghua.edu.cn})}
\thanks{All authors are with the Department of Mechanical Engineering, State Key Laboratory of Tribology in Advanced Equipment, Beijing Key Laboratory of Transformative High-end Manufacturing Equipment and Technology, Tsinghua University,Beijing 100084, China.}%
\thanks{This paper has a supplementary video.}
}

\begin{document}

\maketitle
\thispagestyle{empty}
\pagestyle{empty}

\begin{abstract}

	In this work, we propose a high-voltage, high-frequency control circuit for the untethered applications of dielectric elastomer actuators (DEAs). The circuit board leverages low-voltage resistive components connected in series to control voltages of up to 1.8\,kV within a compact size, suitable for frequencies ranging from 0 to 1\,kHz. A single-channel control board weighs only 2.5\,g. We tested the performance of the control circuit under different load conditions and power supplies. Based on this control circuit, along with a commercial miniature high-voltage power converter, we construct an untethered crawling robot driven by a cylindrical DEA. The 42-g untethered robots successfully obtained crawling locomotion on a bench and within a pipeline at a driving frequency of 15\,Hz, while simultaneously transmitting real-time video data via an onboard camera and antenna. Our work provides a practical way to use low-voltage control electronics to achieve the untethered driving of DEAs, and therefore portable and wearable devices.
\end{abstract}

\section{INTRODUCTION}

Dielectric elastomer actuators (DEAs) are a type of soft actuators that can deform upon the application of an electric field and are among the most promising actuators for soft robots\cite{pelrine2000highspeed,xu2023compact}, haptic devices, and wearable devices \cite{leroy2020multimode}. Typically, DEAs operate within a frequency range of 1\,to\,1000\,Hz and require driving voltages between 1\,kV to 10\,kV to obtain large strains \cite{gu2018softa,tang2023review}, and the higher the voltage applied, the larger deformation a DEA generates\cite{hajiesmaili2021dielectric}. Consequently, the high voltage required for the actuation of DEAs presents significant challenges for the development of lightweight, compact, high-bandwidth, high-voltage (HV) power sources and control circuits for untethered and wearable DEAs \cite{kim2025acrobatics}.

The portable HV power source and control circuits currently used for DEAs adopted several approaches. The first approach involves reducing the driving voltage of DEAs to an extremely low voltage, such as below 1\,kV, and therefore allowing the use of miniature electronic devices. For example, Ji et al. created an ultra-thin-film DEA operating at as low as 300\,V driven by a lightweight 780-mg flyback boost circuit capable of generating voltage of 1\,kHz and 480\,V \cite{ji2019autonomousa}. Similarly, Gravert and colleagues designed actuators operating at 900\,V using commercial HV power modules and Metal Oxide Semiconductor Field Effect Transistors (MOSFETs), resulting in a 15.5\,g drive circuit \cite{gravert2024lowvoltage}. Additionally, Ren et al. developed an ultra-light circuit of only 127\,mg that converted a 7.7\,V input to an output of 600\,V and 400\,Hz \cite{ren2023lightweight}. However, the reduction of DEAs' driving voltage usually compromises DEA's energy and power density or dramatically increases manufacturing complexity\cite{hajiesmaili2021dielectric,ji2019autonomousa}. The second approach, instead, does not compromise driving voltage. This approach controls the driving voltage at the low-voltage Direct Current (DC) end and then amplifies the controlled low voltage to the desired high voltage, avoiding the use of high-voltage controllers \cite{carpi2012electroactive, cao2018untethered}. This approach is suitable primarily for low-frequency applications (\textless\,5\,Hz). For example, Li et al. achieved an output voltage of 3\,Hz and 10\,kV output using a series of boost and flyback converters along with a voltage doubler \cite{li2021selfpowered}. A third approach leverages components that can stand high voltage (e.g., \textgreater\,1\,kV) at the high-voltage end, which typically involves larger device sizes \cite{marette2017full, yoder2024hexagonal}. Schlatter et al. used high-voltage optocouplers to achieve outputs ranging from 1\,kV to 5\,kV, with a maximum frequency of 1\,kHz and a weight of 60\,g \cite{schlatter2018petapicovoltron}. Similarly, Minaminosono and colleagues used high-voltage insulated gate bipolar transistors (IGBTs) to deliver a 2\,kV output with no active discharge circuit, resulting in a limited driving frequency of only 1\,Hz \cite{minaminosono2021untethered}. Oscillatory mechanisms can be constructed to drive DEAs without the use of electronic components \cite{henke2017softa}. Additionally, theoretical calculations have been conducted to explore the charge transfer between DEAs for actuation \cite{mottet2018electric}.

In this paper, we present a high-frequency, high-voltage control circuit for the untethered applications of DEAs. The circuit board leverages low-voltage resistive components connected in series to control voltages of up to 1.8\,kV within a compact size, suitable for frequencies ranging from 0 to 1\,kHz (unloaded), while demonstrating a slew rate of \qty{9.14}{\volt\per\micro\second} under the tested power supply conditions. A single-channel control board weighs only 2.5\,g. We tested the performance of the control circuit under different load conditions and power supplies. Based on this control circuit, along with a commercial miniature DC-HVDC converter, we construct an untethered crawling robot driven by a cylindrical DEA with an equivalent capacitance of 49\,nF, a length of 50\,mm, a diameter of 14\,mm. The 42-g untethered robots successfully obtained locomotion on a bench and within a pipeline at a driving frequency of 15\,Hz, while simultaneously transmitting real-time video data via an onboard camera and antenna. Our work provides a practical way to use low-voltage control electronics to achieve high-voltage, high-frequency driving of DEAs, and therefore portable and wearable devices.

\section{CIRCUIT DESIGN AND PERFORMANCE TESTING}
This study was inspired by the excellent work of Gravert and colleagues on surveying on the footprint of high-voltage MOSFETs in Digi-Key electronics \cite{gravert2024lowvoltage}. Their survey demonstrates that as the blocking drain-source voltage (\(V_{\text{DS}}\)) of MOSFETs increases, their physical sizes and masses grow exponentially. For example, a type of MOSFET modeled as IPN95R3K7P7 from Infineon Technologies, weighing 0.11\,g, a \(V_{\text{DS}}\) of 950\,V and a maximum current of 2\,A, is much smaller and lighter than a higher-voltage one such as the model WSTP3N150 from STMicroelectronics, which has a \(V_{\text{DS}}\) of 1500\,V and a maximum current of 2.5\,A but weighs 2.11\,g --- approximately 20 times that of the former. This comparison indicates that even when two IPN95R3K7P7 units are used in series to accommodate a circuit exceeding 1.5\,kV, their combined volume and mass are still remarkably smaller than that of a single high-voltage component. The key challenge here is how to double the withstand voltage using multiple low-voltage units \cite{alves2024advanced}.

\subsection{Circuit Design and Component Selection}
\begin{figure}[!htb]
	\centering
	\includegraphics[width=0.9\columnwidth]{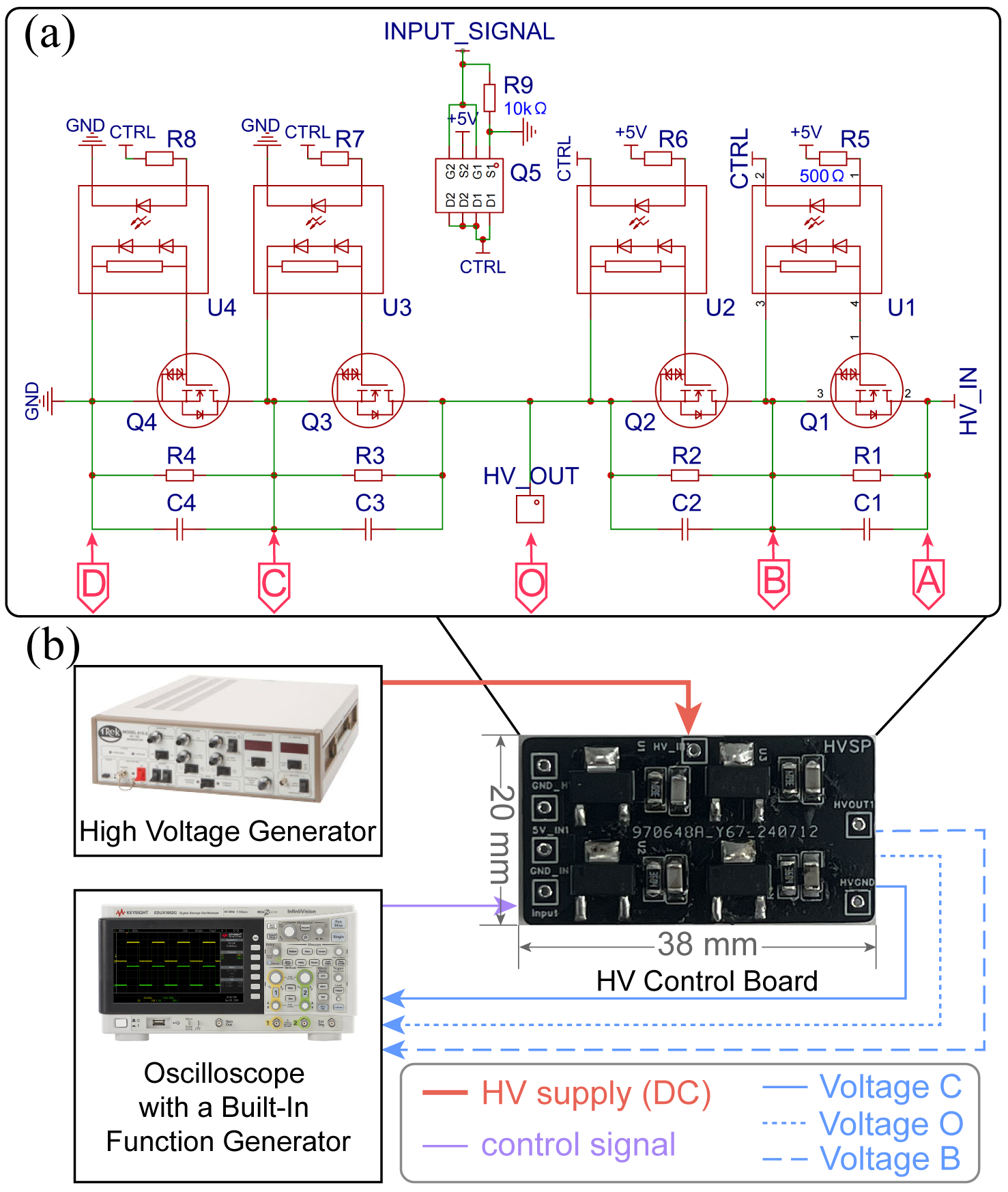}
	\caption{High-voltage control circuit design. (a) Schematic diagram of the HV control circuit. (b) Test setup for the HV control circuit with a benchtop HV generator and an oscilloscope with a built-in function generator.}
	\label{fig:fig1}
\end{figure}

As shown in Fig. \ref{fig:fig1}a, we designed a half-bridge circuit to control the on-off of a high-voltage source. On each side of the bridge, two MOSFETs (blocking \(V_\text{DS}=\text{950}\,\text{V}\)) were connected in series and controlled in synchronization to facilitate the two-state switch. Taking into account the safety guarantees and fault tolerance, we set each MOSFET to sustain up to 900\,V voltage and achieved the control of a 1.8\,kV power source. Each MOSFET was driven by a photovoltaic driver (APV1121SX, Panasonic), with the control signals interconnected to synchronize. The turn-on time and turn-off time for the photovoltaic drivers were 0.4\,ms and 0.1\,ms, respectively. A detailed bill of materials is provided in Table \ref{table:boom}.

\begin{table}[!htb]
    \caption{Bill of Materials for the HV Control Circuit} 
    \centering 
    \begin{tabular}{lllll} 
    \toprule
    C. No. & Description & Part No. & Qty & Wt(g) \\ 
    \midrule
    U1-U4 & MOSFET Driver & APV1121SX & 4 & 0.08 \\ 
    Q1-Q4 & HV MOSFET & IPN95R3K7P7 & 4 & 0.11 \\ 
    Q5 & MOSFET & IRF7509PbF & 1 & 0.02 \\
    C1-C4 & 220 pF & CC1206JKNPODBN221 & 4 & 0.03 \\
    R1-R4 & 3.6 \ensuremath{\text{M}\Omega} & HR1206F3M60P05Z & 4 & \textless\,0.01 \\
    R5-R8 & 500 \ensuremath{\Omega} & ARG05FTC5000 & 4 & \textless\,0.01 \\
    R9 & 10 \ensuremath{\text{k}\Omega} & RMCF0603JT10K0 & 1 & \textless\,0.01 \\
    PCB & 38\,mm \ensuremath{\times} 20\,mm  & Customized & 1 & 1.17 \\
	\midrule
    \multicolumn{4}{r}{Total Weight (After Assembly):} & 2.50\,g \\ 
    \bottomrule
    \end{tabular}
    \label{table:boom} 
    \par \vspace{0.1cm} 
    \raggedright \footnotesize{\parindent=1.2em \indent C. No.: Component Number, Qty: Quantity, Wt: Weight, PCB: Printed Circuit Board.}
\end{table}

\subsection{Voltage Distribution on MOSFETs Connected in Series}

When multiple equivalent MOSFETs were arranged in series, voltages across them may not be equally distributed due to variations in their manufacturing and materials. This uneven distribution of voltages may lead to excessive voltage on MOSFETs with lower leakage currents, causing potential breakdown and subsequent damage to the remaining components. Therefore, the performance tests of the board are crucial to the verification of the proposed method. Electrics tests were conducted using the setup shown in Fig. \ref{fig:fig1}b. The high voltage source was provided by a benchtop HV generator (Model 615-3, Trek) outputting a DC voltage. This DC voltage was then converted into a square-wave AC voltage by our designed HV control board. The voltage of several key points on the board (points A, B, C, D, O in Fig. 1a) was monitored using an oscilloscope (EDUX1002G, Keysight) along with a high-voltage probe (RP1300H, Rigol, 100:1, \qty{1.2}{ns} rise time, \qty{100}{\mega\ohm} input resistance, \qty{5.5}{pF} input capacitance) to characterize the voltage distribution across the series-connected MOSFETs. For safety considerations, we started the tests from a relatively input low voltage (\(V_{\text{HV-in}}\)) of 800\,V. Point A was always as the input voltage and point D was always as the ground thus always zero. Therefore, we emphasized the voltage monitor of points B, C, and O.

\begin{figure}[!htb]
	\centering
	\includegraphics[width=0.9\columnwidth]{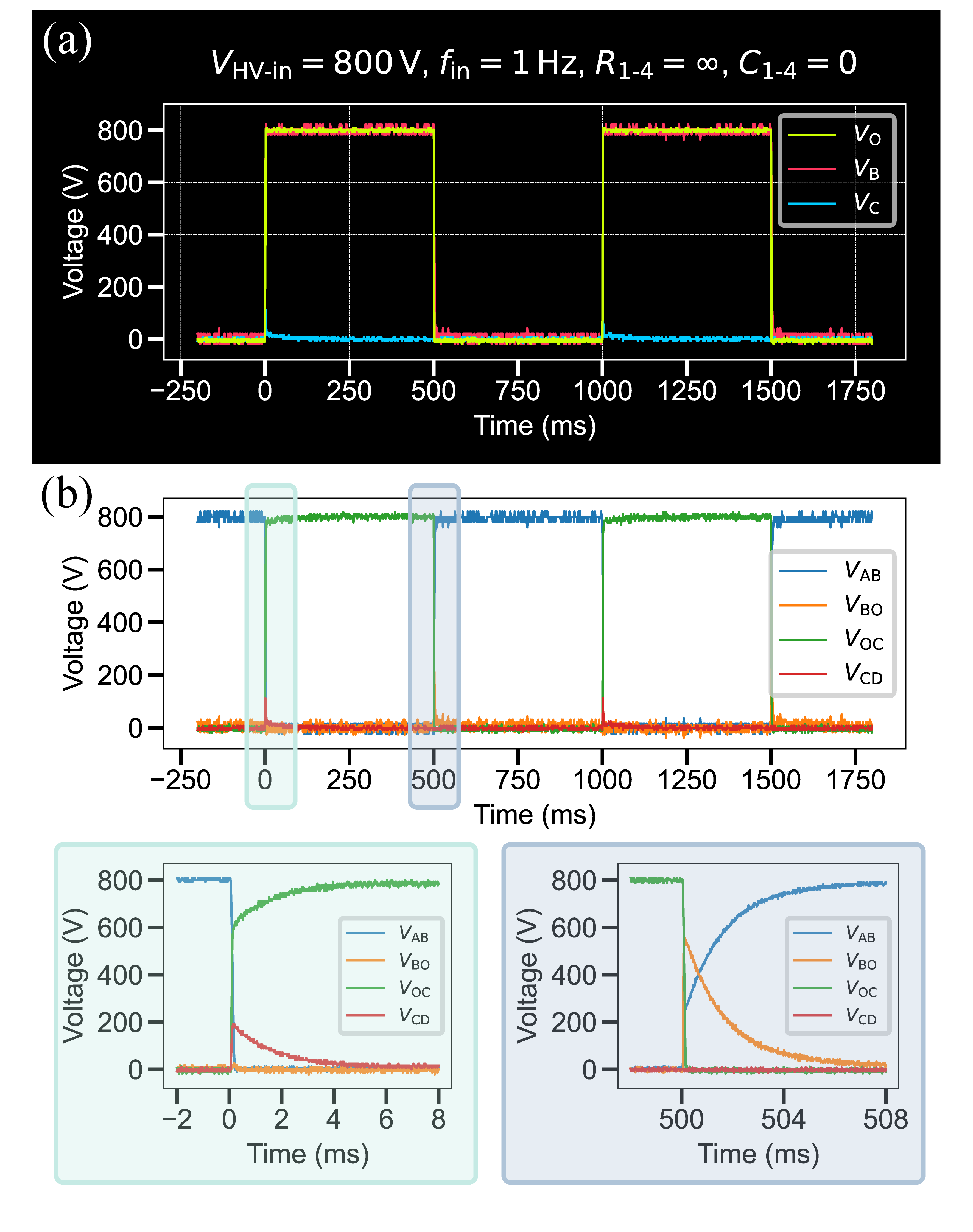}
	\caption{Voltage distribution on MOSFETs with a 1-Hz control signal and 800 V DC power source with no parallel resistors. (a) Direct Voltage measurements at points O, B, and C. (b) Calculate voltage drop on each MOSFET (subfigures on the bottom: zoon-in voltage evolution in the time scale).}
	\label{fig:fig2}
\end{figure}
\begin{figure}[!htb]
	\centering
	\includegraphics[width=0.9\columnwidth]{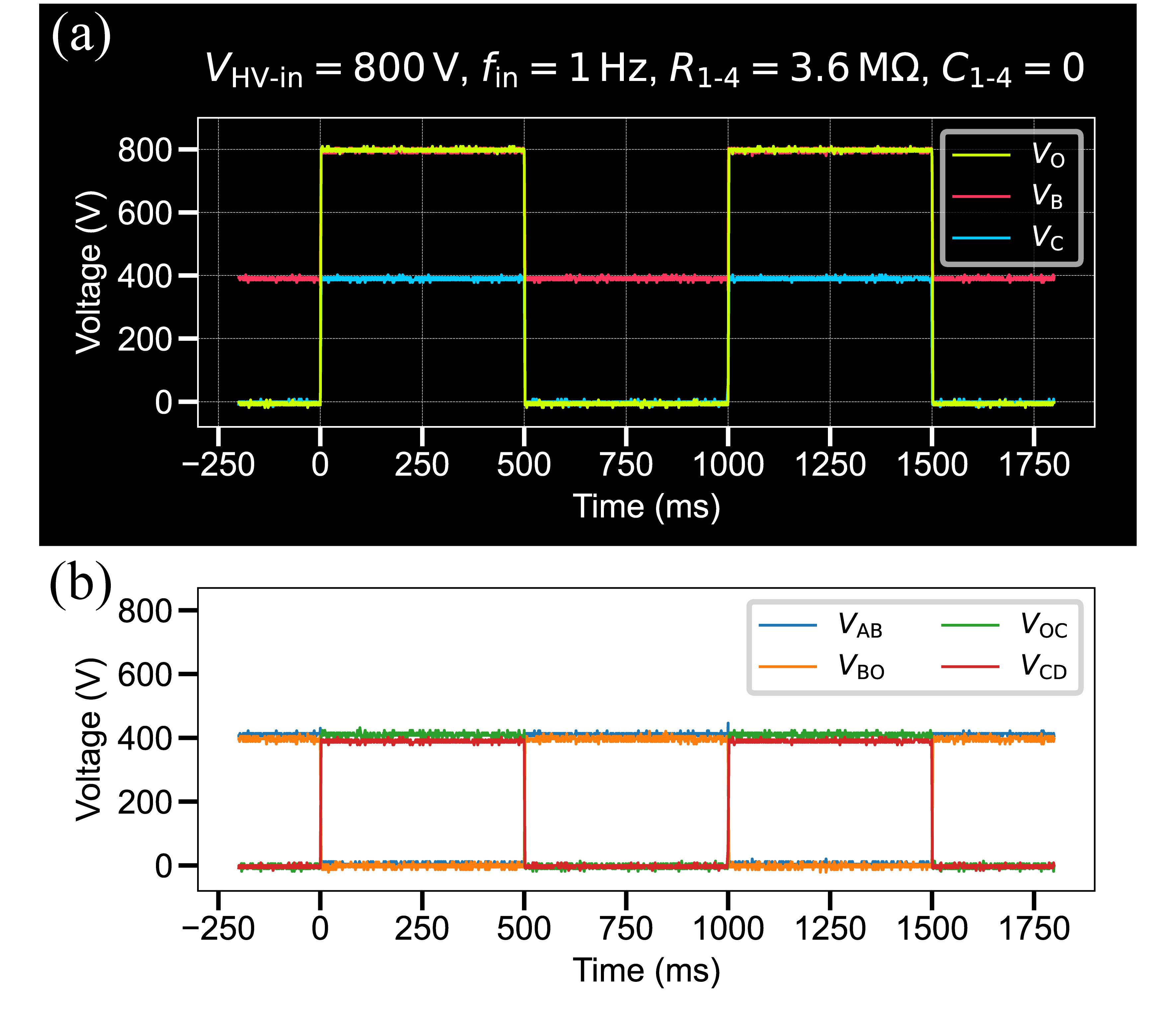}
	\caption{Voltage distribution on MOSFETs with 1-Hz control signal and 800\,V DC power source with 3.6\,\(\mathrm{M\Omega}\) resistors. (a) Direct Voltage measurements at points O, B, and C. (b) Calculate voltage drop on each MOSFET.}
	\label{fig:fig3}
\end{figure}

We first tested the circuit’s performance in converting the DC source power into a low-frequency AC signal. Fig. \ref{fig:fig2}a displayed the voltages at points O, B, and C directed measured by the HV probe and the oscilloscope in refer to the ground (\(V_\text{A}\), \(V_\text{B}\), and \(V_\text{C}\)) with the control frequency (\(f_{\text{in}}\)) set to 1\,Hz. Fig. \ref{fig:fig2}b illustrated the calculated voltage drop each of the four MOSFETs (\(V_\text{AB}\), \(V_\text{BO}\), \(V_\text{OC}\), and \(V_\text{CD}\)). The lower two subfigures are the zoom-in voltage evolution of the rising and falling edges in the time scale. The experimental results indicate that without paralleled resistors across the MOSFETs, the voltage was unevenly distributed to a single MOSFET on each side, and the other almost had no voltage crossing over, failing to meet our requirements. Conversely, paralleling large resistors across the MOSFETs resulted in a more uniform voltage distribution with minimal error, satisfying our needs. The parallel resistors should be selected as much smaller than the output resistors of the MOSFETs. However, too small parallel resistors caused leakage current and thus amplitude reduction in the output voltage. To compromise the two effects, we chose a resistor of 3.6\,\(\mathrm{M\Omega}\) paralleled across each MOSFET to aid in voltage distribution. The measured voltage and the calculated voltage drop on each MOSFET are shown in Fig. \ref{fig:fig3}a and Fig. \ref{fig:fig3}b. The voltages were equally distributed on the two MOSFETs on each side, verifying our major consumption.

\subsection{Voltage Distribution on MOSFETs with High-Frequency Control Signals}

\begin{figure}[!htb]
	\centering
	\includegraphics[width=0.9\columnwidth]{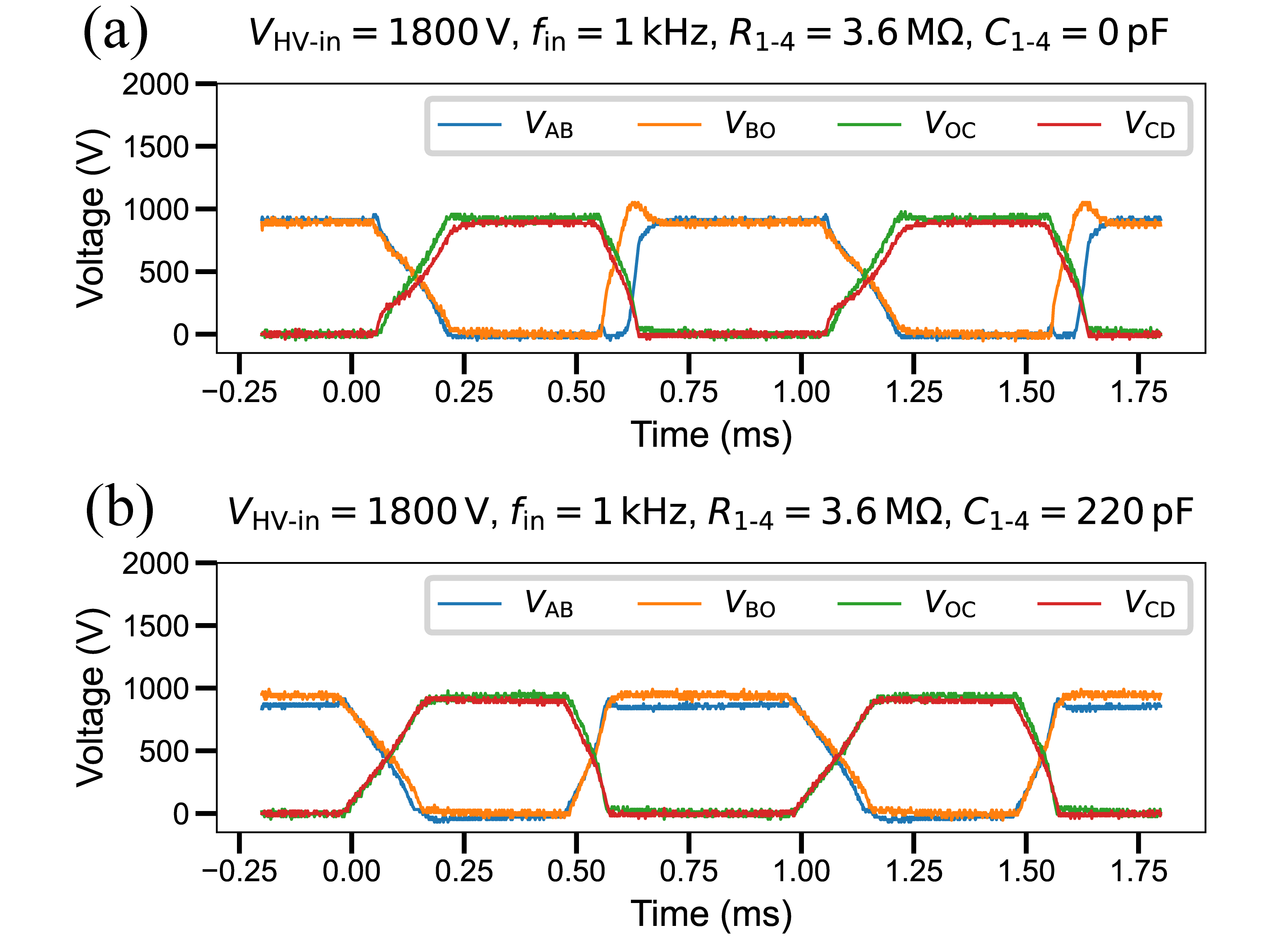}
	\caption{Voltage distribution on MOSFETs with a 1-kHz control signal and 1800 V DC power source with 3.6-\(\mathrm{M\Omega}\) resistors. (a) Calculated voltage drop on each MOSFET with no capacitor across. (b) Calculated voltage drop on each MOSFET with capacitors across.}
	\label{fig:fig4}
\end{figure}

We further tested the voltage distribution on the MOSFETs with high-frequency control signals. Due to the switching delays of the photovoltaic MOSFET driver and the response delays of the MOSFETs, the series-connected MOSFETs exhibited uneven voltage distributions during transient state control. As shown in Fig. \ref{fig:fig4}a, with \(V_{\text{HV-in}} = 1.8\,\text{kV}\) and \(f_{\text{in}} = 1\,\text{kHz}\), \(V_{\text{BO}}\) and \(V_{\text{AB}}\) demonstrated an uneven voltage distribution at the moment of switch-off, leading to potential breakthrough on the MOSFET with higher voltage drop. Furthermore, the test results indicate that the control circuit achieves a slew rate of \qty{9.14}{\volt\per\micro\second}, demonstrating its capability to operate effectively at \qty{1}{\kilo\hertz}. 
In Fig. \ref{fig:fig4}b, a 220\,pF capacitor was paralleled across each MOSFET to limit the switch-off time, thereby absorbing the impact and ensuring a uniform voltage distribution during the transition.

\subsection{Test of the control circuit on a simulated DEA load}

\begin{figure}[!b]
	\centering
	\includegraphics[width=0.9\columnwidth]{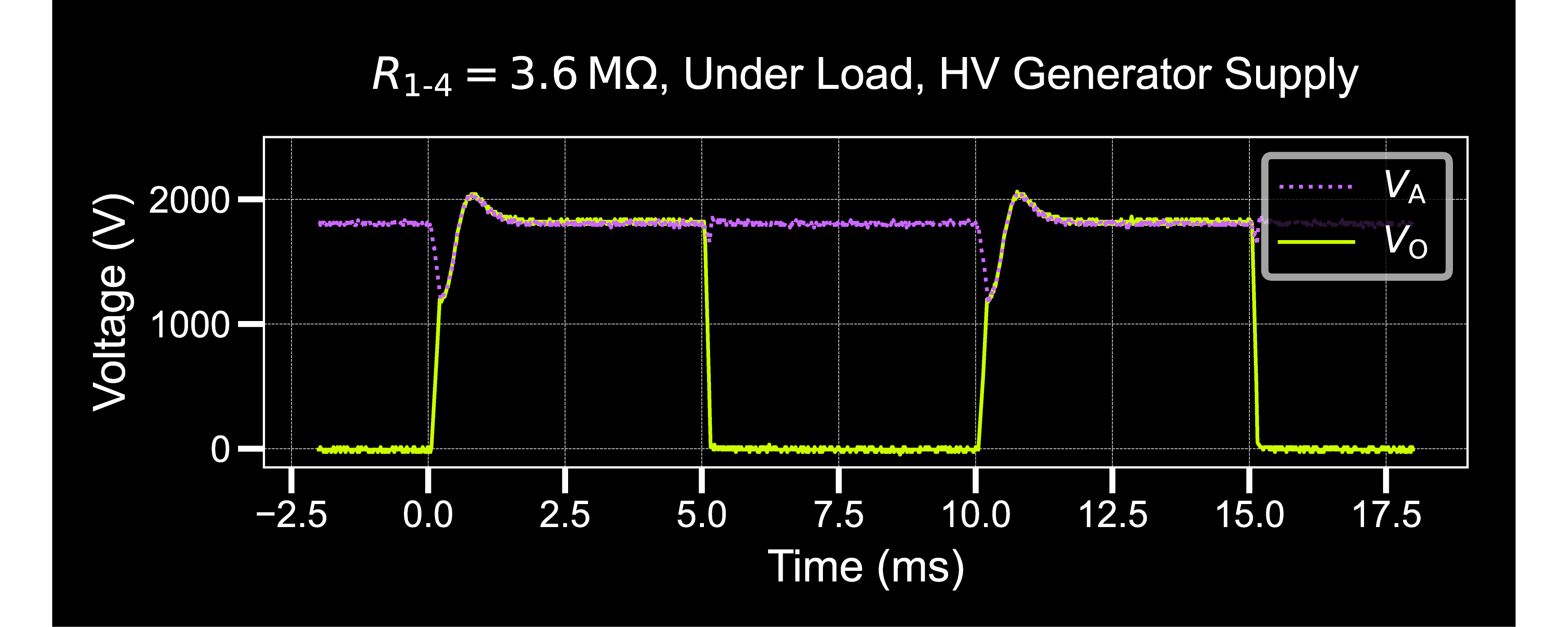}
	\caption{Test of the control circuit on a simulated DEA circuit with a 10\,nF capacitor and 100\,M\(\mathrm{\Omega}\) resistor at a 100\,Hz control signal and a 1.8\,kV source voltage.}
	\label{fig:fig5}
\end{figure}

A millimeter- to centimeter-size DEA typically exhibits an equivalent capacitance of 500\,pF to 50\,nF, an equivalent serial resistance (\(R_\text{s}\)) of 10\,k\(\mathrm{\Omega}\) to 100\,k\(\mathrm{\Omega}\), and an equivalent parallel resistors (\(R_\text{p}\)) exceeding 10\,M\(\mathrm{\Omega}\) \cite{duduta2019realizing,shi2022processable,zhao2018compacta}. In our tests, a 100\,k\(\mathrm{\Omega}\) resistor(RN3WS100KFT/BA1, Tyohm) and a 10\,nF capacitor (Y5V103M2KV16CC0204, KNSCHA) were connected in series to mimic the electric load from a DEA for assessing the performance of the HV control board. Fig. \ref{fig:fig5} displays the voltage of the input point (\(V_\text{A}\)) and the output point load voltage \(V_O\) at a control frequency of 100\,Hz and a DC source voltage of 1.8\,kV (these are typical frequency and voltage for driving a DEA). The results indicate that there are some fluctuations in \(V_\text{O}\) and \(V_\text{A}\) caused by the HV generator but the switch speed and amplitude were as designed, demonstrating the practical potential of our circuit for driving a DEA.

\section{Performance Test of the Control Circuit with a Miniature Power Supply}\label{sec:testing}
\subsection{Simulation and Characterization of Power Supply from the Miniature DC-HVDC Converter}
\begin{figure}[!htb]
	\centering
	\includegraphics[width=0.9\columnwidth]{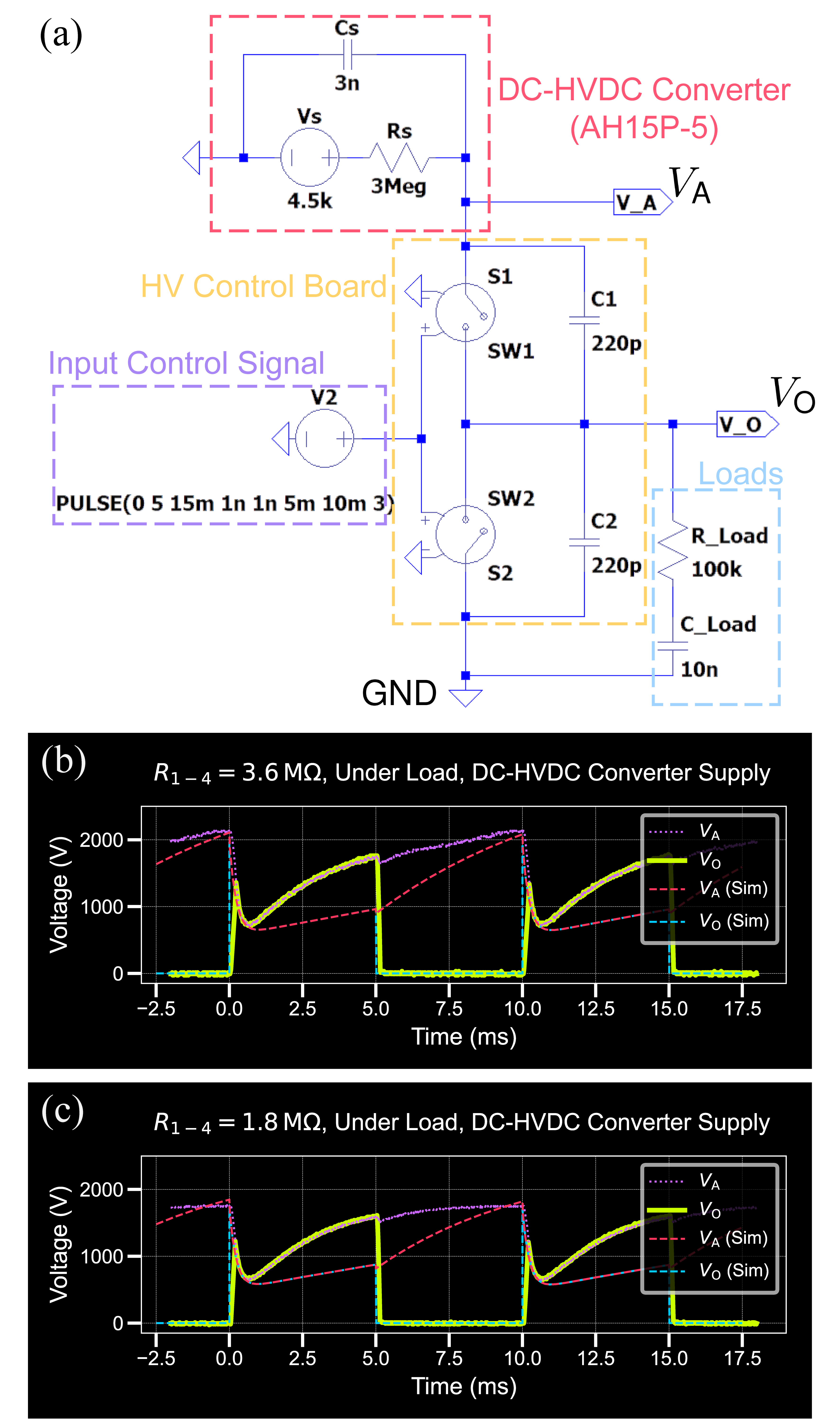}
	\caption{The dynamic response of the control circuit with the power source from a miniature DC-HVDC converter supply. (a) Circuit model of the miniature DC-HVDC converter and the HV control board in LTspice. (b) Simulation results with 3.6\,M\(\mathrm{\Omega}\) paralleled resistors at 100\,Hz. (c) Experimental results with 1.8\,M\(\mathrm{\Omega}\) paralleled resistors at 100\,Hz.}
	\label{fig:fig6}
\end{figure}

To achieve untethered integration and meet the voltage requirements of a DEA, we need not only the portable control board but also a miniature HV power supply. In this part, we tested the performance of our designed control circuit in combination with a commercial miniature DC-HVDC converter (AH15P-5, XP Power), which weighs 3.86\,g and has a rated output power of 1.5\,W and a rated output voltage of 1.5\,kV. According to its datasheet, the rated output voltage is only achieved when the external resistance equals the internal resistance, and the no-load output voltage can reach up to 4.5\,kV. Based on its datasheet and the dynamic response observed during tests, we modeled the converter as a non-ideal voltage source with an internal resistor of 3\,M\(\mathrm{\Omega}\), a supply voltage of 4.5\,kV, and a parallel capacitance of 3\,nF, as illustrated in Fig. \ref{fig:fig6}a. The MOSFETs on the HV control board were modeled as ideal voltage-controlled switches with an on-resistance, as the switching frequency is relatively low (100 Hz) and the dominant circuit behavior can be captured without considering parasitic effects. The simulation was conducted in LTspice software to simulate the circuit's dynamic response. The simulation data and actual experimental results at a control frequency of 100\,Hz are shown in Figs. \ref{fig:fig6}b and \ref{fig:fig6}c. In Fig. \ref{fig:fig6}b, the paralleled resistor \(R_\text{1-4}\) of the each MOSFET was 3.6\,M\(\mathrm{\Omega}\). During the discharge cycle, the source voltage could exceed 1.8\,kV, potentially leading to the breakdown of the MOSFETs in the high-voltage side of the half-bridge circuit. Therefore, the resistors \(R_\text{1-4}\) in Fig. \ref{fig:fig1}a were replaced with two 3.6\,M\(\mathrm{\Omega}\) resistors in parallel, equivalent to 1.8\,M\(\mathrm{\Omega}\), as shown in Fig. \ref{fig:fig6}c, to ensure that the voltage throughout the cycle did not exceed the breakdown voltage. We attribute the discrepancies between the simulation data and the actual measurements to the nonlinearity of the actual load and the simplifications made in the modeling of the DC-HVDC converter and the MOSFETs. Though with some differences in simulation and experiment, we successfully demonstrated that with a miniature HV DC power supply, we were able to control a 1.8\,kV power source with MOSFETs connected in series. This combination of miniature power supply with our proposed portable control circuit will be sufficient for driving DEAs without tethers.

\subsection{Test of the Untethered HV Control Circuit with Different External Loads and Frequencies}\label{subsec:evaluation}
\begin{figure}[!htb]
	\centering
	\includegraphics[width=0.9\columnwidth]{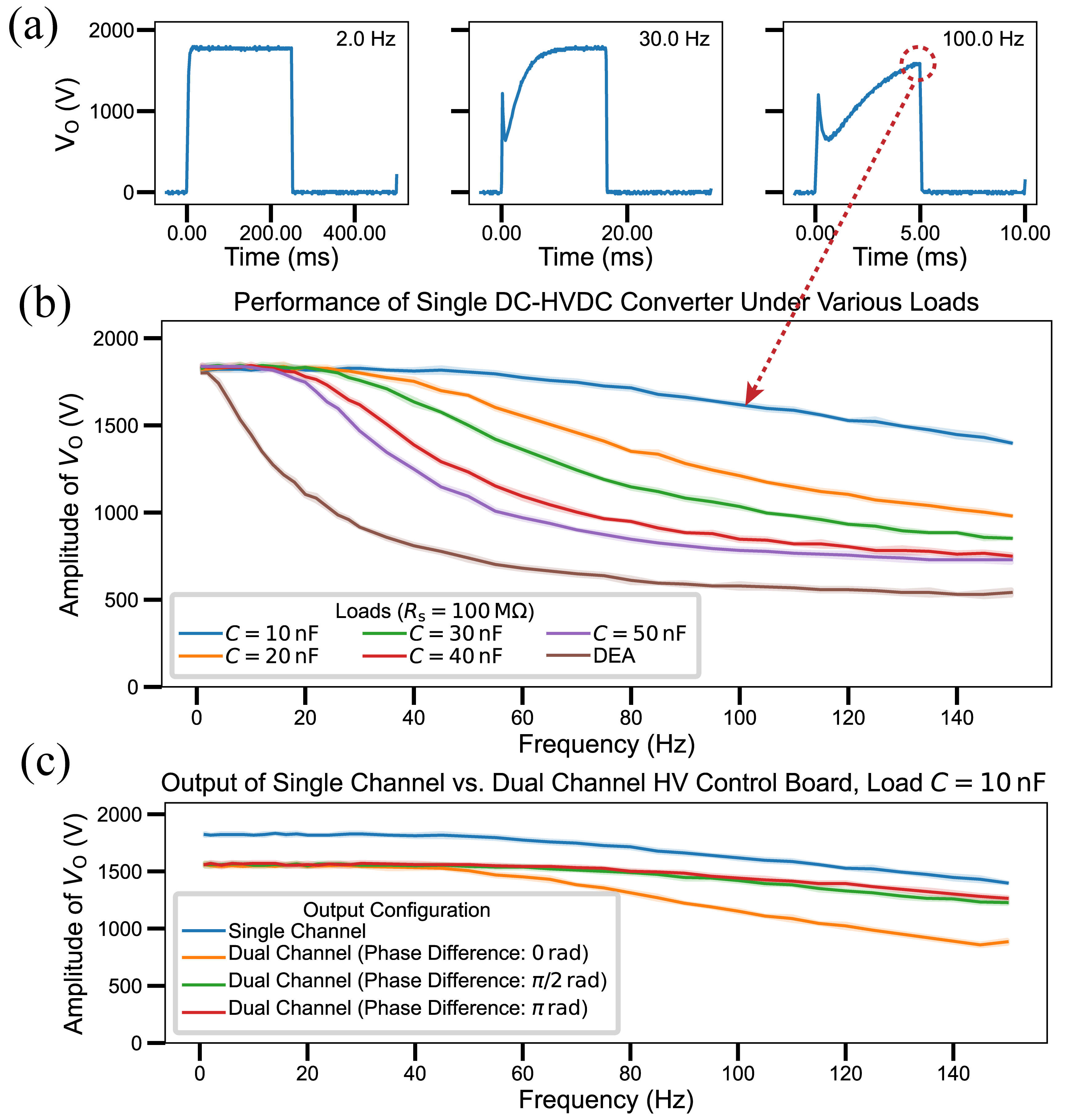}
	\caption{Experimental Results of the HV control circuit with the DC-HVDC converter with different external loads and frequencies. (a) The circuit's voltage waveforms with a 10\,nF load at 2\,Hz, 30\,Hz, and 100\,Hz. (b) Voltage amplitude response over frequency for different capacitive loads. (d) Performance of dual-channel outputs with 3 phase differences.}
	\label{fig:fig7}
\end{figure}

The untethered HV control system consists of the miniature DC-HVDC converter and our proposed control circuit. The maximum output current from the miniature converter was only 0.66\,mA, which may limit the performance of the system. A multilayered DEA with 5 layer of electrodes and 6 layers of dielectrics (Elkem Silbione\textsuperscript{TM} LSR 4305 and Dow SYLGARD\textsuperscript{TM} 184 at a 3:1 mass ratio) was fabricated for testing \cite{shao2025long}, each layer having a thickness of 50\,\textmu m. This multilayered film was rolled into a cylindrical structure with a length of 50\,mm and a diameter of 14\,mm. We performed system identification of the DEA at 10\,Hz and 1\,kV, which showed that the DEA was equivalent to a capacitance of 49\,nF, a serial resistance \(R_\text{p}\) of 60\,k\(\mathrm{\Omega}\), and a parallel resistance of 6.6\,M\(\mathrm{\Omega}\). To test the output capability of the control circuit at different frequencies and external loading conditions, besides the DEA, we further constructed various simulated loads ranging from 10\,nF to 50\,nF serially connecting with a 100-k\(\mathrm{\Omega}\) resistor. Fig. \ref{fig:fig7}a displays the voltage waveforms with a 10\,nF load at 2\,Hz, 30\,Hz, and 100\,Hz, all with a set power source as 1.8\,kV. The maximum output voltage under different loads and frequencies were plotted in Fig. \ref{fig:fig7}b, which indicated that larger capacitive loads caused faster voltage decays over frequencies due to the miniature converter's limited output power. Additionally, real DEA actuators exhibited greater decay than the simulated loads using capacitors, mainly because used the capacitances of ceramic capacitors decreased at high voltages \cite{coday2018characterization}, whereas the DEA experienced a small increase in capacitance due to induced strain \cite{jiang2022long}.

\subsection{Performance Testing for Dual Outputs}

Additionally, we tested the performance of a single DC-HVDC converter connected to two units of the HV control circuit for dual-channel output, with each's load as 10\,nF. The experimental setup also examined the phase differences of 0, \(\pi/2\), and \(\pi\) between the two output channels. The results, displayed in Fig. \ref{fig:fig7}c, were as expectations that the attenuation from the dual-channel output was slightly lower at high frequencies compared with a single channel. Moreover, when the channels operated with a phase difference, the peak power requirement was reduced so that the output performance was enhanced. The dual-channel output experiment shows potential that our circuit can be expanded into a larger scale in the future, for driving multi-actuator systems.

\section{Test of the Performance of a DEA and an Untethered Mobile Robot with the Control Circuit}

Based on the DEAs fabricated in Sec. \ref{subsec:evaluation}, this section integrates all components (the control circuit, the DC-HVDC converter, and a battery) to achieve locomotion for a DEA-driven mobile robot, enabling the DEA to operate with everything onboard in a compact volume \cite{pei2002multifunctional,shao2022untethered,rich2018untethereda}.

\subsection{Characterization of a DEA's Displacement with a Benchtop Power Supply and a Miniature Converter}
\begin{figure}[!htb]
	\centering
	\includegraphics[width=0.9\columnwidth]{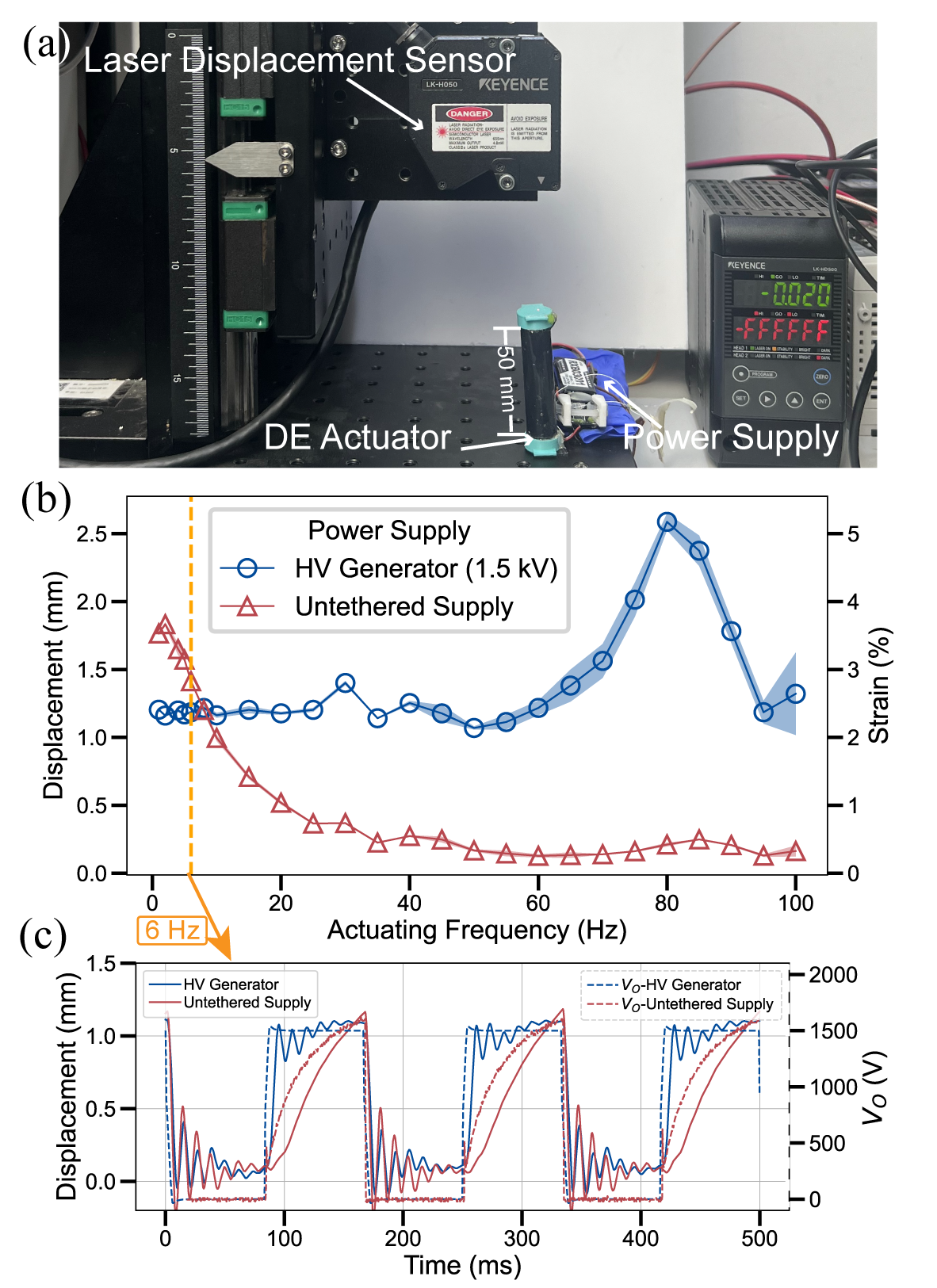}
	\caption{Displacement characterization of DEA under different power supplies and driving frequencies. (a) Experimental setup for displacement measurement of DEA. (b) Displacement amplitude across different frequencies. (c) Time response of the DEA's displacement and circuit's output voltage at 6\,Hz.}
	\label{fig:fig8}
\end{figure}
The free displacement of the cylindrical DEA at changing frequencies generated by the control circuit was characterized. As shown in Fig. \ref{fig:fig8}a, one end of the DEA was fixed, and the displacement at the other end was measured using a laser displacement sensor (LK-H050, KEYENCE). Two power supplies were compared: an benchtop HV generator and the miniature DC-HVDC converter tested in Sec. \ref{sec:testing}, which will be referred to as the untethered supply in later sections. The Part 1 in the supplementary video gives an intuitive comparison of the two power supplies. The DEA's displacement amplitude at different driving frequencies is depicted in Fig. \ref{fig:fig8}b. It's observed that at low frequencies (e.g., \textless 10\,Hz), the displacement from the HV generator and the untethered supply were comparable. However, at higher frequencies, due to the miniature converter's power limitations, the displacement driven by the untethered supply was much smaller than the HV generator. To further verify the attenuation in displacement for the untethered power supply was due to limited power, in Fig. \ref{fig:fig8}c, we give the displacement and output voltage change over time in a few cycles at 6-Hz driving frequency. This figure clearly shows that the charging speed driven by the untethered supply was much slower compared to the HV generator, and thus exhibited a delay that significantly increased the electrical time constant of the system. This limitation indicates that for generating higher power output, a higher-power converter is required, which usually means larger size for the converter.

\subsection{Untethered Miniature Robots Driven by Dielectric Elastomer Actuators}

\begin{figure}[!htb]
	\centering
	\includegraphics[width=0.9\columnwidth]{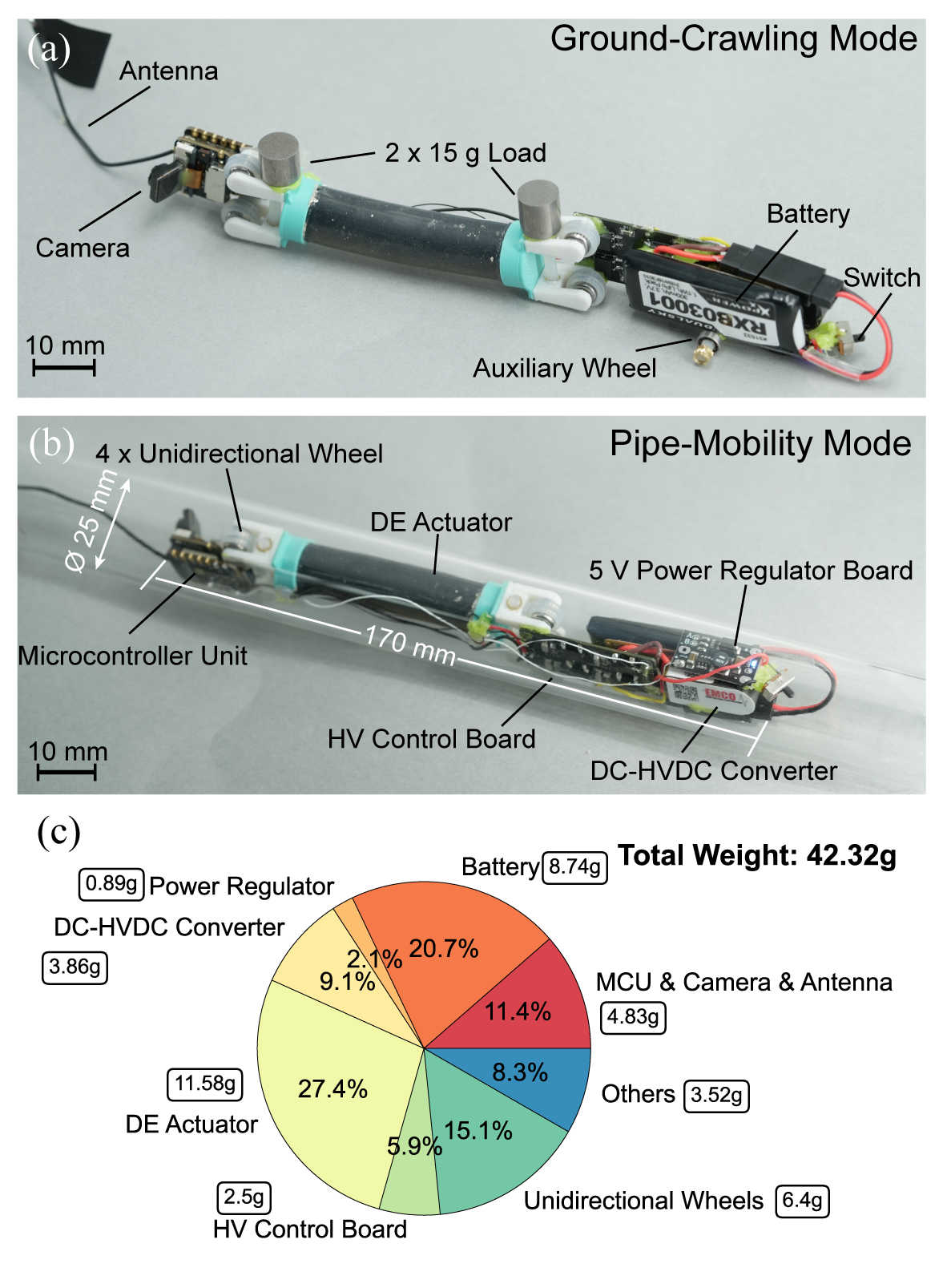}
	\caption{Design and integration of the untethered mobile robot. (a) Ground-crawling mode. (b) Pipe-mobility mode. (c) Mass distribution of different functional components of the robot.}
	\label{fig:fig9}
\end{figure}

All components required for a mobile robot were integrated in an untethered locomotion robot as shown in Fig. \ref{fig:fig9}. There included a lithium battery (RXB03001 3.7\,V 1.1\,Wh, DUALSKY) and a power regulator to set the battery's output at 5\,V. The battery powered the DC-HVDC converter in Section 3 to generate a DC high voltage, and our proposed HV control board was used to produce the high-frequency high-voltage square waves for driving the DEA. The driving signals were controlled by a microcontroller unit (ESP32S3, Seeed), and a camera with a resolution of 1600\(\times\)1200 (OV2640, Seeed) was incorporated to transmit image signals in real-time via a 2.4\,G WiFi. The mechanical structure of the robot included two needle one-way clutch bearings (HF0306, XIFANG) at each end of the DEA, coupled with silicone rubber non-slip rings to form unidirectional wheels. The total length of the robot was 170\,mm, with a radial dimension of 25\,mm. It featured two locomotion modes: with an auxiliary wheel to maintain balance and the addition of two loads at the front and rear ends to adjust the mass distribution, it could crawl on the ground as shown in Fig. \ref{fig:fig9}a; it was also capable of moving within a 25-mm diameter pipe as shown in Fig. 9b without the auxiliary wheel. Fig. \ref{fig:fig9}c presents the robot's mass percentage of different functional components. The total mass of the robot was 42.3\,g, with the DEA and battery being the heaviest components, accounting for 27.4\% and 20.7\% respectively. The DC-HVDC converter and HV control board contributing 9.1\% and 5.9\%, respectively.

\begin{figure}[!htb]
	\centering
	\includegraphics[width=0.9\columnwidth]{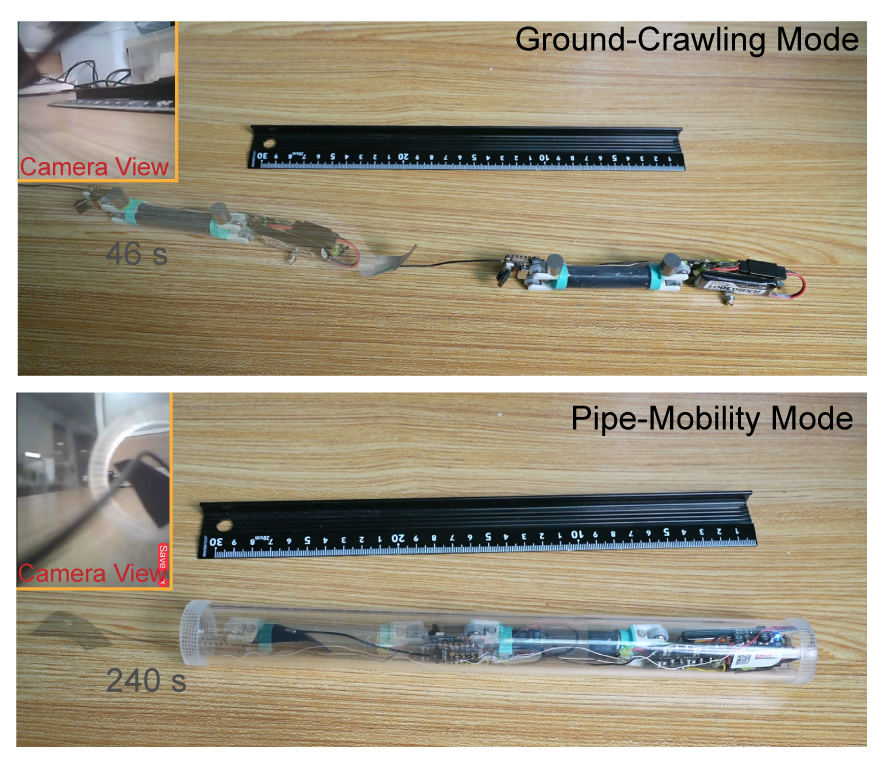}
	\caption{Movement demonstration of the untethered mobile robot. (a) Ground-crawling mode movement. (b) Pipe-mobility mode movement.}
	\label{fig:fig10}
\end{figure}

Finally, we conducted motion tests on the untethered mobile robot. The robot successfully obtained the expected untethered motion in two modes and could simultaneously transmit real-time video from the camera's perspective. As depicted in Fig. \ref{fig:fig10}a, in the ground-crawling mode, the robot moved approximately 30\,cm over 46\,s, and in Fig. \ref{fig:fig10}b in the pipe-mobility mode, due to the more complex interactions between the robot and the environment, it moved approximately 15\,cm within 240\,s. The Part 2 in the Supplementary Video gives more details of the locomotion demonstration.

\section{CONCLUSIONS}


In this study, we designed and developed a high-frequency high-voltage control circuit building untethered DEA-driven robots or systems. The proposed design, utilizing multiple low-voltage resistive components arranged in series, achieved a compact and lightweight solution capable of controlling voltages up to 1.8\,kV at frequencies ranging from 0 to 1\,kHz. The performance of the control circuit was evaluated through comprehensive testing under different load conditions and power supplies. Furthermore, the successful integration of the high-voltage control board and a miniature converter into an untethered DEA-driven mobile robot showcased its potential for real-world applications. 

The major contribution of this work lies in proposing a compact, lightweight, and low-cost power control solution for untethered DEA-driven devices. By enabling the untethered operation of DEAs, this study opens up new possibilities for the application of DEAs in soft robotics, haptic devices, and wearable devices, where free-space locomotion and portability are crucial. However, the performance of the robot in pipe-mobility mode can be further improved through structural design optimizations. Future work will focus on further optimizing the circuit design as well as the DEA to enhance their power efficiency and locomotion capabilities.





\bibliographystyle{IEEEtran}
\bibliography{ref}




\end{document}